\documentclass[a4paper]{./svproc}
\usepackage{url}
\newcommand{\etal}{\textit{et al.}} 
\usepackage{amsfonts}
\usepackage{amsmath}
\usepackage{amssymb}
\usepackage{color,soul}
\usepackage{graphicx}
\usepackage{multirow} 
\usepackage{booktabs} 

\usepackage{prettyref}
\newrefformat{fig}{Fig.~\ref{#1}}
\newrefformat{sec}{Sec.~\ref{#1}}
\newrefformat{tab}{Table~\ref{#1}}
\newrefformat{algo}{Algo.~\ref{#1}}
\newrefformat{eq}{(\ref{#1})}
\newrefformat{prob}{\textit{Problem~\ref{#1}}}

\begin{document}
\mainmatter              
\title{A Haptic-Based Proximity Sensing System for Buried Object in Granular Material}
\titlerunning{Proximity Sensing in Granules}  
%
\author{Zeqing Zhang\inst{1,2,}\thanks{This work was done when he was an intern at Baidu Research, Baidu, Beijing, China.} \and Ruixing Jia\inst{1}, Youcan Yan\inst{3,}\thanks{Corresponding author.} \and Ruihua Han\inst{1} \and Shijie Lin\inst{1} \and \\ Qian Jiang\inst{4} \and Liangjun Zhang\inst{5} \and Jia Pan\inst{1,2}}
\authorrunning{Z. Zhang et al.} 
%
\tocauthor{Zeqing Zhang, Ruixing Jia, Youcan Yan, Ruihua Han, Shijie Lin, Qian Jiang, Liangjun Zhang, and Jia Pan
}
\institute{The University of Hong Kong, Hong Kong, China,\\
\email{\{zzqing, ruixing, hanrh, lsj2048\}@connect.hku.hk, jpan@cs.hku.hk}
\and
Center for Transformative Garment Production, Hong Kong, China,
\and
CNRS-University of Montpellier, Montpellier, France,\\
\email{youcan.yan@lirmm.fr}
\and
The Hong Kong Polytechnic University, Hong Kong, China,\\
\email{qian2020.jiang@connect.polyu.hk}
\and
Robotics and Autonomous Driving Lab of Baidu Research, Sunnyvale, USA.\\
\email{liangjunzhang@baidu.com}
}

\maketitle              

\begin{abstract}
The proximity perception of objects in granular materials is significant, especially for applications like minesweeping. However, due to particles' opacity and complex properties, existing proximity sensors suffer from high costs from sophisticated hardware and high user-cost from unintuitive results. In this paper, we propose a simple yet effective proximity sensing system for underground stuff based on the haptic feedback of the sensor-granules interaction. We study and employ the unique characteristic of particles -- failure wedge zone, and combine the machine learning method -- Gaussian process regression, to identify the force signal changes induced by the proximity of objects, so as to achieve near-field perception. Furthermore, we design a novel trajectory to control the probe searching in granules for a wide range of perception. Also, our proximity sensing system can adaptively determine optimal parameters for robustness operation in different particles. Experiments demonstrate our system can perceive underground objects over $0.5$ to $7$ cm in advance among various materials.

\end{abstract}
\section{Introduction}
Granular materials (GMs) are common in everyday life, such as sand in nature, soil in civil engineering, and grain in agriculture. They can be defined as a group of solid particles with amorphous shapes and complicated properties, such as jamming \cite{richard2005slow,liu2010jamming}, fluidization \cite{zik1992mobility}, dilatancy \cite{onoda1990random}, and bifurcation \cite{gravish2010force}. As such, GMs cannot simply be classified as either solids or liquids \cite{jaeger1996granular}. 
In some practical applications, like excavation, minesweeping, and archaeology, unknown buried objects beneath GMs need to be handled with great care due to their fragility, danger, or vulnerability. Therefore, it is desired to develop a proximity sensing system that can sense the presence of the buried object in advance without direct contact. 

Currently, commercial proximity sensors mostly work in the air, such as the ultrasonic transducer for the car's blind-spot monitors and the infrared sensor for contactless sinks. 
Unlike the air with almost no effect on signal propagation, the liquid will have a larger effect on the perception modality. Therefore, most of the underwater proximity sensors use sonar \cite{hover2007vehicle}, a small part uses vision \cite{girdhar2023curee}, and a few part is based on flow field information \cite{liu2022underwater}. 
However, when it comes to the granule scenario, there are even fewer options available, primarily due to the opaqueness and complex physical properties inherent in such media. Proximity sensors for buried utilities under granules can be divided into intrusive and non-intrusive ways, depending on whether the sensor is in contact with particles. 
The ground penetrating radar \cite{daniels2004ground} is a widely utilized sensor for non-intrusive proximity sensing, employing electromagnetic waves to detect underground objects based on variations of electrical conductivity in media. However, its operation and interpretation necessitate the expertise of professionals \cite{travassos2020artificial}. Another type of non-intrusive proximity sensor is based on sound waves, e.g., \cite{frazier2000acoustic,aranchuk2023laser,sugimoto2011buried}. Nevertheless, these devices are either complex or bulky in size. So, some researchers have started to investigate lightweight proximity sensors in an invasive manner directly using the interaction force signal. A recent work \cite{jia2022autonomous} reports the use of a pre-defined model to pre-touch subterranean obstacles. However, due to the fixed rupture distance, this kind of sensing is strictly confined to a particular GM and a set of manual tuning parameters. Also, due to the complex characteristics of particles, the analytical interaction haptic model can not be obtained. Therefore, the development of an intrusive proximity sensor that can work in a variety of  GMs still presents significant challenges.

\begin{figure}[tbp]
    \centering
    \includegraphics[width=0.450\textwidth]{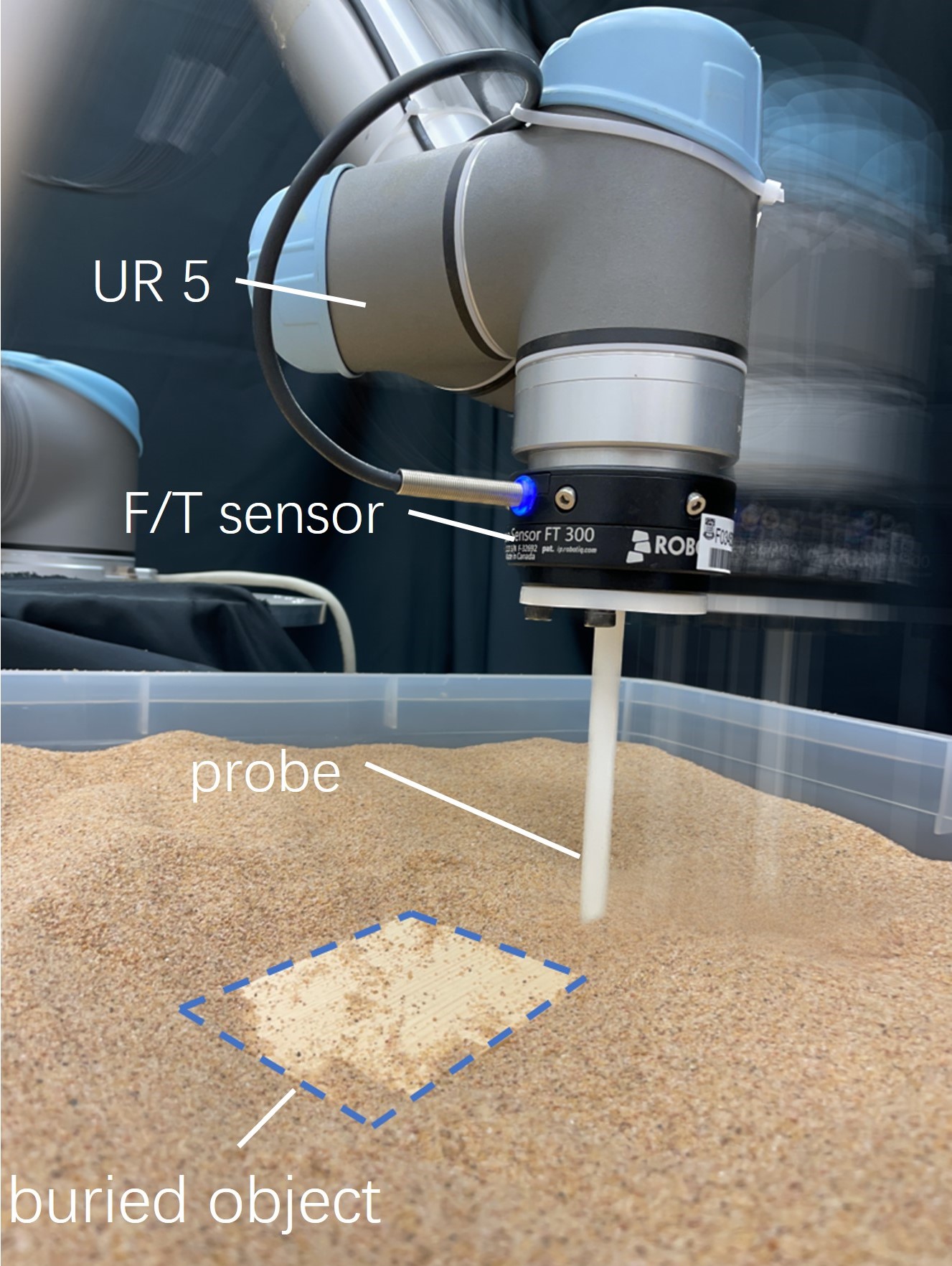}
    \vspace{-10pt}
    \caption{Prototype of our haptic-based proximity sensing system, GRAINS, for the near-field perception of the buried object in granules.}
    \label{fig:prototype}
    \vspace{-10pt}
\end{figure}

In this paper, we propose a simple yet effective real-time near-field sensing system, called Granular-Material-embedded Autonomous Proximity Sensing System (GRAINS), as shown in \prettyref{fig:prototype}. It is an intrusive sensor that uses the failure wedge zone \cite{swick1988model} ahead of the probe generated when particles interact with the probe as an airbag layer to contact the buried object in advance. The proximity of underground objects is determined according to the difference of haptic feedback in the unique jamming state \cite{richard2005slow,liu2010jamming} of particles during the interaction between the failure wedge zone and objects. So, the probe-objects collision can be avoided. Specifically, the main contributions of this work can be summarized as follows.
\vspace{-5pt}
\begin{itemize}
    \item[1] We propose an intrusive proximity sensing system, GRAINS, based on the online learning of interaction force signals between the probe and granules.
    \item[2] We discover that the failure wedge zone can work as the airbag of the probe in GMs to avoid the probe-objects collision.
    \item[3] We control the failure wedge zone by designing the probe motion and also investigate the haptic signal for proximity sensing using interaction force feedback between the failure wedge zone and subsurface objects.
    \item[4] We present an autonomous parameter calibration process, allowing the system to adaptively update optimal parameters and work robustly in different particles.
\end{itemize}

\vspace{-10pt}

\section{Related Work}
In this section, we will review the proximity sensors that are commonly used by industry and proposed by researchers to detect buried objects in GMs. They are categorized as non-intrusive and intrusive methods. 

As a non-intrusive method, ground penetrating radar \cite{daniels2004ground} is a common tool in construction and geophysical applications. It transmits electromagnetic radiation and detects underground stuff by interpreting reflected waves. Even if it could be a compact system and provide robust results at meter level, it suffers from massive post-processing work \cite{travassos2020artificial} and requires qualified people to operate it. Similar devices, like EZiDIG \cite{EZiDIG2}, can be mounted on the excavator to increase excavation confidence. However, it only responds to metallic objects in the soil such as pipes and cables, and also requires operators to be trained in advance. The use-cost of these devices increases as a result. In addition, based on acoustics, \cite{frazier2000acoustic} provides a sensing system using a set of complicated sonar arrays toward the ground. Due to different principles, the aforementioned sensors allow meter-level perception, but a group of laser-acoustic sensors, e.g., \cite{aranchuk2023laser,sugimoto2011buried}, only feel the presence of buried utilities within several centimeters. Normally, these systems are complicated and bulky, since they mainly contain acoustic sources, the laser generator, the high power supply, and other accessory equipment.

In comparison, intrusive sensing systems are quite simple and lightweight. They mainly employ unique granular properties, even if GMs have complicated natures. For instance, several studies primarily focus on the haptic modality. Syrymova \etal \cite{syrymova2020vibro} introduce a vibro-tactile method that classifies the absence or presence of a rigid body in GM based on mechanical vibrations by squeezing granules in a rubber balloon. Utilizing the vision-based tactile sensor, GelSight \cite{yuan2017gelsight}, the Digger Finger in \cite{patel2021digger} estimates the shape of simple objects in GM by penetrating into GM and acquiring touch feedback. Jia \etal \cite{jia2017multimodal,jia2021tactile} use multiple tactile modalities from the haptic sensor BioTac (SynTouch LLC, Los Angeles, CA) to estimate contact states between robotic fingertips and objects in GM. Obviously, these tactile techniques require direct contact between the sensory surface and buried objects, and can not sense obstacles in advance. Most related work on proximity sensing in GM comes from  \cite{jia2022autonomous}, which is based on the characteristic of force variation between tool and object and employs a pre-defined model to estimate the distribution of underground stuff automatically. It can be found this work is highly constrained to a specific GM since lots of manual tuning parameters are used in the sensor model based on preliminary experiments and empirical equations. For other granules, those parameters may not work anymore. 
In this work, we also use the force of the probe interacting with the particles as a proximity signal. However, the Gaussian process regression (GPR) \cite{rasmussen2003gaussian} is employed to learn the recent historical data, and then predict a short period of future data to compare whether the force pattern is normal or not. So, our method can be greatly applied to different types of particles without any pre-defined parameters and can constantly learn the complicated interaction model even in one type of granule considering the complex granule properties.

\section{Problem Statement}\label{sec:problem_statement}

We outline a region filled with single homogeneous granules to be searched by the sensor as a two-dimensional (2D) area $\mathcal{W}_g = \{\mathbf{x} \in \mathbb{R}^{2} | (x, y)\}$. We assume that one or more objects $ \mathcal{O}_i = \{ \mathbf{x} \in \mathbb{R}^{2} | (x, y) \}\ (i = 1, \cdots, N) $ are buried in this region, i.e., $\mathcal{O}_i \subset \mathcal{W}_g, \forall i$. We do not make assumptions about the height of the granule surface. But we only consider the horizontal proximity perception in this work, thus, we assume that the depth of the subsurface object must be within the depth range that the sensor can perceive. 

During the exploration of the proximity sensor through particles, a set of online measurements 
\begin{equation}\label{eq:observation}
    \mathcal{T} = \{(\mathbf{x}_t, \mathbf{f}_t) | t = 1, \cdots, L\}
\end{equation}
are obtained, where $\mathbf{x}_t \in \mathcal{W}_g$ is the position of the sensor, and $\mathbf{f}_t = [f^t_x, f^t_y]$ is the interaction force readings in $x,\ y$ axes measured from the load cell. Our goal is to employ the observation $\mathcal{T}$ to determine the presence of the underground object nearby and then stop the exploration at time $t = L$ to avoid the potential collision. 
So, in this work, we represent the proximity perception problem as follows.
\begin{problem}\label{prob:proxi_prob}
    Let $\xi$ be a function $\xi: \mathcal{T} \rightarrow \{0, 1\}$ to determine whether buried objects are nearby. The \textbf{proximity perception problem} is to find this function such that
\begin{equation}
    \begin{array}{rcl}
      &\xi(\mathcal{T}) =
      \left\{
        \begin{array}{rcl}
        1 &, & \ \ when\ an\ object\ gets\ close,\\
        0 &, & \ \ others. \\        
        \end{array} \right. \\     
    \end{array}
\end{equation}
\end{problem}
%
The proximity degree can be evaluated by the given metric.
\begin{definition}
    The \textbf{proximity sensing range} is defined by a function $\zeta: \xi \rightarrow \mathbb{R}$ to measure the minimum distance between the buried objects and the position that the sensor perceives one of them, i.e.,
    \begin{equation}\label{eq:sensing_range}
        \zeta(\xi) = \min || \mathbf{x}_o - \mathbf{x}_L ||_2,
    \end{equation}
    where $\mathbf{x}_o \in \mathcal{O}_i, \forall i$ and $\mathbf{x}_L \in \mathcal{T}|_{t = L}$.
\end{definition}

\section{Physical Principle}
\label{sec:principle}
\begin{figure}[htbp]
    \centering
    \includegraphics[width=0.95\textwidth]{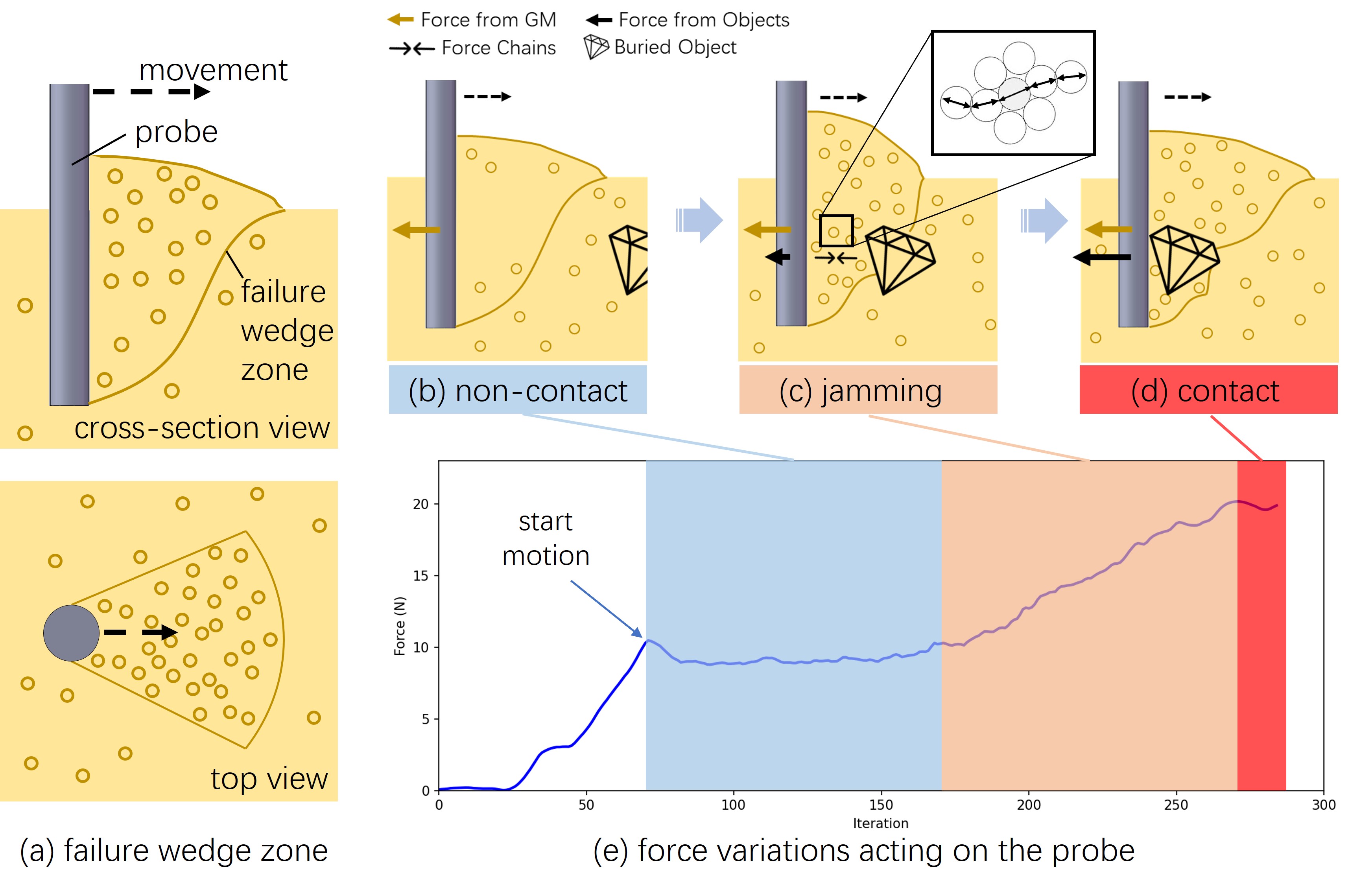}
    \vspace{-15pt}
    \caption{Physical principle. (a) Failure wedge zone ahead of the probe as it moves in granules. (b)-(d) State changes of the failure wedge zone as the probe approaches the object in GMs and (e) the corresponding force variations. Inset: force chains (from \cite{peters2005characterization}) when the granule jamming occurs. }
    \label{fig:fwzone_states_forces}
    \vspace{-15pt}
\end{figure}

In this section, we will explain the principle of our proximity sensing system GRAINS and reveal how to find the near-field perception function $\xi$ based on the haptic feedback $\mathbf{f}_t$ in the online measurement $\mathcal{T}$. 
The key of our proximity perception method for buried objects in GMs lies in the use of the failure wedge zone \cite{swick1988model}, which is the unique characteristic of granules. The failure wedge zone forms ahead of the probe, when a probe is dragged through a homogeneous granular medium, as shown in \prettyref{fig:fwzone_states_forces}-(a). This area appears wedge-shaped from the cross-section view and fanned from the top view. Experiments show that, as the probe moves, particles entering the failure wedge zone will be perturbed into motion, while particles outside this region remain still. 
%
To avoid contact between the tool and buried objects, we can consider the failure wedge zone in front of the tool as the airbag layer. So, we can avoid the probe-object collision by using the contact between the failure wedge zone and the object.

Let's look at the state changes of the failure wedge zone as the probe approaches the object in GMs, as depicted in \prettyref{fig:fwzone_states_forces}-(b) to (d). The haptic feedback acting on the probe at each state is also illustrated in \prettyref{fig:fwzone_states_forces}-(e).
\begin{itemize}
    \item \textbf{Non-contact State}: When the probe is distant from the buried object, it experiences only resistive forces from particles within the failure wedge zone, which is roughly constant \cite{albert1999slow}. See blue area in \prettyref{fig:fwzone_states_forces}-(e). 
    \item  \textbf{Jamming State}: When the object enters the failure wedge zone, the granule jamming occurs. Then, besides forces from particles, the resistive force from the object can be added to the probe (orange region in \prettyref{fig:fwzone_states_forces}-(e)) through force chains \cite{peters2005characterization} (the inset of \prettyref{fig:fwzone_states_forces}-(c)) among squeezed granules.
    \item  \textbf{Contact State}: The probe contacts the object directly, which is what we don't want.
\end{itemize}
According to haptic variations during above three states, we can find that the interaction between the failure wedge zone and rigid object can lead to force signals reliable for near-field sensing. 
Therefore, the \textbf{principle} of the proposed proximity sensor GRAINS is to recognize the jamming state by using the changes of force signal $\mathbf{f}_t | _{t = L}$ from measurement $\mathcal{T}$ during the granule jamming, so as to stop the movement of the tool before the contact state and avoid the collision at position $\mathbf{x}_t | _{t = L}$.


\section{Sensor Design}
\subsection{Prototype}
The prototype of GRAINS is simple and compact, as explicated in \prettyref{fig:prototype}. It is composed of three parts, i.e., the 6-DoF industrial robotic manipulator (UR5, Universal Robots, Denmark), the 6-axis F/T sensor (Force Torque Sensor FT 300, Robotiq, Canada), and a 3D-printed rod with $1$ cm diameter and $14$ cm in length. 
As proof of concept, the robot arm UR5 used here is only to automatically control the probe motion and easily acquire its position $\mathbf{x}_t$. In the future version, the probe can be driven by a motor or other mechanism and even installed on a mobile platform using other positioning devices. Similarly, the current F/T sensor could be replaced with a low-cost tactile sensor in the future, as long as the measurements $\mathcal{T} = \{(\mathbf{x}_t, \mathbf{f}_t)\}$ could be obtained simultaneously.

\subsection{Motion Control}\label{sec:motion_control}
\begin{figure}[tbp]
    \centering
    \includegraphics[width=0.60\textwidth]{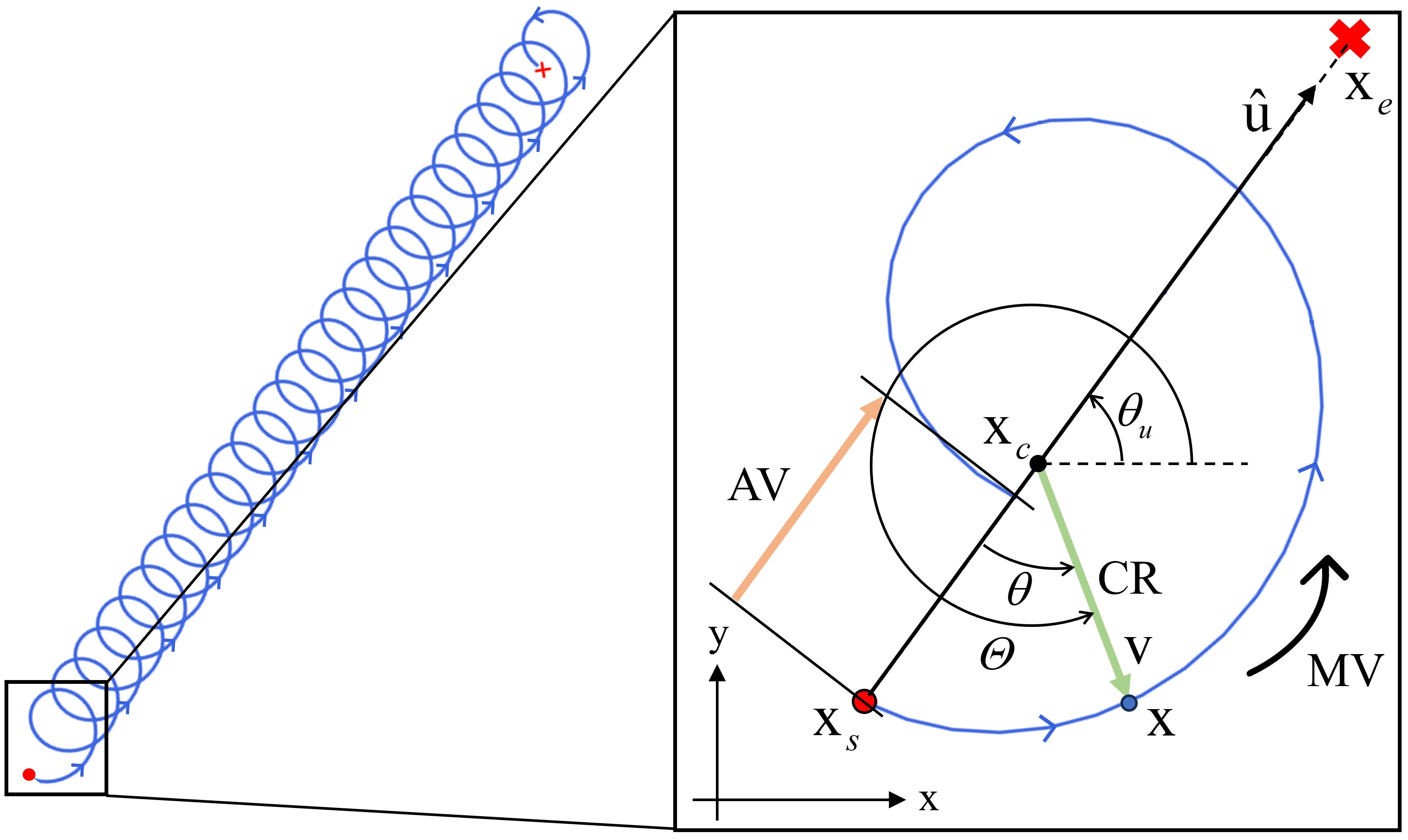}
    \vspace{-10pt}
    \caption{Definition of the spiral trajectory. Enlarged view: one circular motion in the spiral trajectory, i.e., $\theta \in [0, 2\pi]$.}
    \label{fig:def_spiral_traj}
    \vspace{-15pt}
\end{figure}

Due to the fan-shaped distribution (from top-view) of the failure wedge zone ahead of the probe, only the buried object directly in front of the probe's motion can compress GMs, resulting in a jamming state. So, the objects located on two sides of the movement cannot be detected by the probe because they do not enter the failure wedge zone. To enlarge the perception area, we design a novel trajectory to control the probe searching objects in all directions. We call this motion as \textbf{sprial trajecotry}, combining the linear advancement and circular
vibration, where the linear motion ensures forward exploration and circular
vibration rotates the failure wedge zone. As illustrated in \prettyref{fig:def_spiral_traj}, the spiral trajectory is defined by three parameters, that is, the circular radius (CR), advance velocity (AV), and motion velocity (MV). Here CR (unit: m) refers to the radius of circular movement, and AV (unit: m) indicates the forward distance of the probe after one circular motion. From \prettyref{fig:def_spiral_traj}, it can be found that CR and AV fully determine the path of spiral movement, and MV is a dimensionless variable from UR5 controller, ranging from 0 to 1, presenting the ratio of its end-effector speed. 

The path of the spiral trajectory can be formulated as follows. Given start and end positions $\mathbf{x}_s$ and $\mathbf{x}_e$, the vector from the starting point to goal can be expressed as
\begin{equation}
    \mathbf{u} = \mathbf{x}_e - \mathbf{x}_s \triangleq [x_u, y_u]^\intercal
\end{equation}
The unit vector of $\mathbf{u}$ and its angle could be calculated by
$\hat{\mathbf{u}} = {\mathbf{u}}/{||\mathbf{u}||_2}$ and $\theta_u = \arctan(y_u/x_u)$.
By definition of the spiral trajectory, the center point $\textbf{x}_c$ of the circular motion will move along the direction of $\hat{\mathbf{u}}$. At the same time, the probe will rotate around this moving center. Let's consider one circular motion in the spiral trajectory for simplicity, as depicted in the enlarged view in \prettyref{fig:def_spiral_traj}. In this case, the probe will advance the distance of AV, which is also the distance of the center $\textbf{x}_c$ after one circular motion. Without loss of generality, we suppose the probe has been rotated $\theta \in [0, 2\pi]$ around the center. Then the distance of the center that has been moved along $\hat{\mathbf{u}}$ can be calculated by
\begin{equation}\label{eq:delta_distance}
    \Delta u_c = \frac{\theta}{2\pi}\text{AV} 
\end{equation}
Thus, the current position of the center is 
\begin{equation}
    \mathbf{x}_c = \mathbf{x}_s + \text{CR} \cdot \hat{\mathbf{u}} + \Delta u_c \cdot \hat{\mathbf{u}} \triangleq [x_c, y_c]^\intercal
\end{equation}
Then, the current position of the probe $\mathbf{x} \triangleq [x,y]^\intercal$ along the spiral trajectory can be formulated by
\begin{align}\label{eq:spiral_xy}
    \begin{split}
        x & = x_c + \text{CR}\cdot\cos{\Theta},\\
        y & = y_c + \text{CR}\cdot\sin{\Theta},
    \end{split}    
\end{align}
where $\Theta$ is the angle of vector $\mathbf{v}$ and can be determined according to the geometrical relationship from \prettyref{fig:def_spiral_traj} as:
\begin{equation}\label{eq:global_angle}
    \Theta = \theta_u + \pi + \theta,
\end{equation}
If we uniformly discretize $[0, 2\pi]$ into $H$ segments, then for each angle, i.e., 
\begin{equation}\label{eq:theta_pointwise}
    \theta_i = 0 + i \cdot (2\pi/H),\  i = 0,\cdots,H
\end{equation}
we can generate the pointwise positions $\mathbf{x}_i$ along the spiral trajecotry by substituting \prettyref{eq:theta_pointwise} into \prettyref{eq:delta_distance}-\prettyref{eq:global_angle}, yielding
\begin{equation}\label{eq:path_xi}
    \mathbf{x}_i = \mathbf{x}_s + (\text{CR} + \frac{\theta_i}{2\pi}\text{AV})\cdot \hat{\mathbf{u}} + \text{CR}\cdot\begin{bmatrix}
        \cos (\theta_u + \pi + \theta_i)\\
        \sin (\theta_u + \pi + \theta_i)
    \end{bmatrix}.
\end{equation}
From \prettyref{eq:path_xi}, it can be found that the path of spiral trajectory only relies on CR and AV.
Given the path, we can control the motion velocity of the probe along this path by tuning the dimensionless variable MV.

\begin{figure}[htbp]
    \centering
    \vspace{-15pt}
    \includegraphics[width=0.99\textwidth]{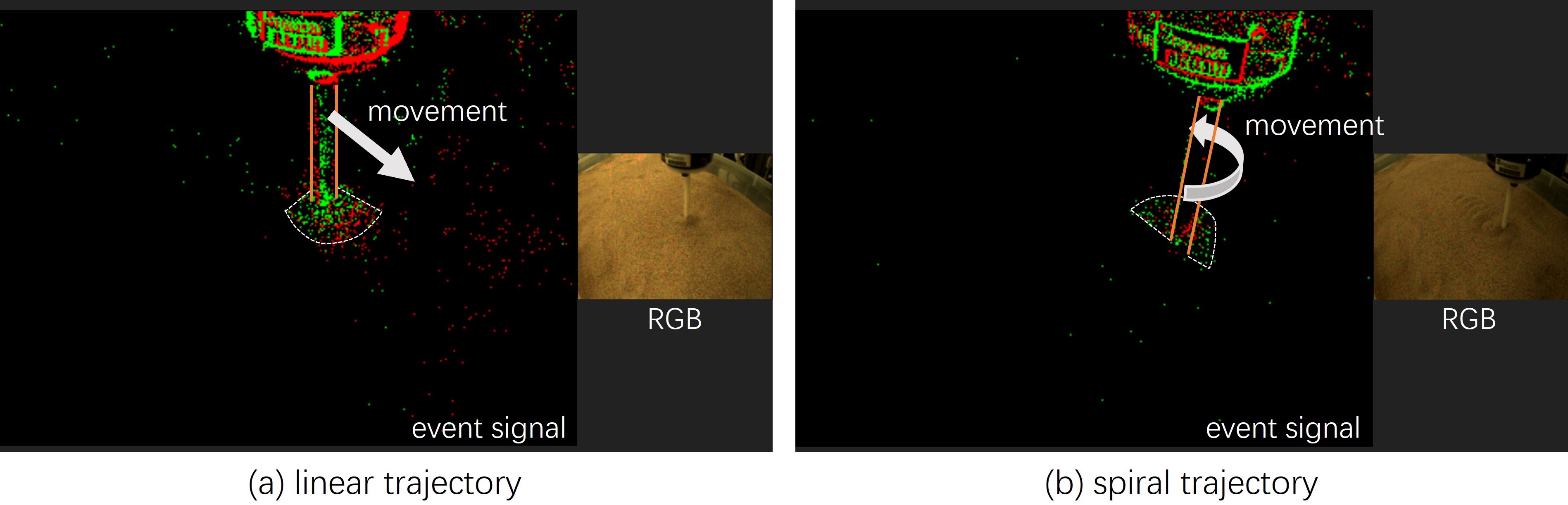}
    \vspace{-10pt}
    \caption{Snapshots of the failure wedge zone ahead of the probe along (a) linear and (b) spiral trajectories.}
    \label{fig:fwzong_vis_evcamera}
    \vspace{-20pt}
\end{figure}

The goal of the spiral trajectory we design is to enlarge the proximity perception area via rotating the failure wedge zone during the exploration in GMs. The following experiment validates our design.
\prettyref{fig:fwzong_vis_evcamera} demonstrates the snapshots of the failure wedge zone ahead of the probe along linear and spiral trajectories, respectively. Note that here we employ
the event camera \cite{gallego2020event}, a dynamic vision sensor with high sensitivity to dynamic motions, to record the probe motion along two trajectories. In simple terms, the event camera is capable of capturing the pixels of moving objects, but it cannot detect stationary objects. Consequently, as illustrated in \prettyref{fig:fwzong_vis_evcamera}-(a), the region highlighted in front of the probe (outlined by the dashed curve) exhibits a fan-shaped zone, that is, the failure wedge zone. From \prettyref{fig:fwzong_vis_evcamera}-(b), the failure wedge zone rotates due to the circular motion involved in spiral trajectory, enabling the probe to detect surrounding objects around $360^{\circ}$.

\subsection{Haptic-Based Proximity Sensing}
According to \prettyref{sec:principle}, we know our proximity sensor perceives the nearby object in advance via the variations in the haptic feedback. However, due to the complicated properties of GMs, the analytical interaction force model can not be obtained, compared to the well-studied contact models for rigid objects and liquids. Prior works primarily focus on the interaction \textit{force value} under different simplifications, such as RFT model \cite{maladen2009undulatory,li2012resistive} or a pre-defined threshold \cite{jia2022autonomous}, to deal with the complex interaction signals.

In this work, we solve this problem by examining the \textit{force pattern} rather than focusing on some discrete force values. 
\begin{definition}
    Here the force pattern $\mathcal{F}$ is defined as a sequence of force magnitudes over a time series, i.e.,
    \begin{equation}\label{eq:force_pattern}
        \mathcal{F} = \{(t, f_t) | f_t = \sqrt{(f^t_x)^2 + (f^t_y)^2}, t = 1, \cdots M\},
    \end{equation}
    where $t$ refers to the time index and $f_t$ is the magnitude of force measurement $\mathbf{f}_t$ in \prettyref{eq:observation} at time $t$.
\end{definition}
Note that $M$ in \prettyref{eq:force_pattern} should be smaller than $L$ in \prettyref{eq:observation}.
Let $\mathbf{t} = [1, \cdots, M] \in \mathbb{R}^{M}$.  We can assume that the force magnitude $f_t$ is related to every $t$ as follows
\begin{equation} \label{eq:f_t_prior}
    f_t = p(t) + \epsilon_t,\ \forall t \in \mathbf{t},
\end{equation}
where $p(t)$ is a Gaussian Process (GP) specified by a mean function and a covariance
function:
\begin{equation}\label{eq:def_GP}
    p(t) \sim \mathcal{GP}(m(t)\equiv0,k(t,t))),
\end{equation}
and $\epsilon_t$ is independently and identically distributed Gaussian noise:
\begin{equation}
    \epsilon_t \sim \mathcal{G}(0, \sigma^2).
\end{equation}
If defining $\mathbf{f} = \{f_1, \cdots, f_M\}$ and $\epsilon = \{\epsilon_1, \cdots, \epsilon_M\}$, then \prettyref{eq:f_t_prior} can be reformulated equivalently as 
\begin{equation}
    \mathbf{f} = p(\mathbf{t}) + \epsilon.
\end{equation}
Given the force pattern $\mathcal{F}$, we can predict the real-value output $f_*$ for a new input index $t_*$ by the Gaussian regression process (GPR) \cite{rasmussen2003gaussian}. Here $\mathcal{F}$ is deemed as the training set. Then, given a set of test inputs $\mathbf{t}_* = [M+1, \cdots, M+M_*]\in \mathbb{R}^{M_*}$, the posterior distribution $p(\mathbf{t}_*)$ conditioned on $\mathcal{F}$ is a Gaussian distribution expressed as
\begin{equation} \label{eq:gpr_posterior}
    \begin{split}
        p(\mathbf{t}_*|\mathcal{F}) \sim \mathcal{GP}(
        &K(\mathbf{t}_*,\mathbf{t})[K(\mathbf{t},\mathbf{t})+\sigma^2I]^{-1}\mathbf{f},\\
        &K(\mathbf{t}_*,\mathbf{t}_*) - K(\mathbf{t}_*,\mathbf{t})[K(\mathbf{t},\mathbf{t})+\sigma^2I]^{-1}K(\mathbf{t},\mathbf{t}_*),
    \end{split}
\end{equation}
where $I$ is the identity matrix, and the matrix $K(\mathbf{t},\mathbf{t}_*)$ is an $M\times M_*$ matrix containing covariances evaluated at all pairs of training and test data, and similarly for other matrices $K(\mathbf{t},\mathbf{t})$, $K(\mathbf{t}_*,\mathbf{t})$, and $K(\mathbf{t}_*,\mathbf{t}_*)$.

In this work, due to the spiral trajectory, the force pattern $\mathcal{F}$ exhibits a periodic pattern in the time domain, which can be modeled using the Exp-Sine-Squared (ESS) kernel (aka periodic kernel) \cite{duvenaud2014automatic}:
\begin{align}\label{eq:periodic_kernel}
    k_{ESS}(t_i,t_j) = \sigma_p^2 \exp \left(-\frac{2}{l^2}\sin^2 \left(\pi\frac{|t_i-t_j|}{T}\right)\right),
\end{align}
where $t_i, t_j \in \mathbf{t}$, and $\sigma_p^2$ is the overall variance, and $l$ is the length scale, and $T$ refers to the periodicity of the kernel.
To account for the high randomness from GMs, the white noise kernel \cite{duvenaud2014automatic} is used to estimate the noise of the measured forces:
\begin{align}\label{eq:noise_kernel}
    k_{white}(t_i,t_j) = \sigma_w^2 I,
\end{align}
where $\sigma_w^2$ is the noise variance and $I$ is the identity matrix. The sum-kernel
\begin{equation}\label{eq:sum_kernel}
    k = k_{ESS} + k_{white}
\end{equation}
is employed as the prior for the covariance function in \prettyref{eq:def_GP}.

Finally, the force magnitude distributions $\mathbf{f}_*$ corresponding to $\mathbf{t}_*$ can be predicted by mean and covariance functions by evaluating \prettyref{eq:gpr_posterior}. For example, the predicted force magnitude $f_*$ at the time $t_* \in \mathbf{t}_*$ can be expressed by the GP distribution, denoted as: 
\begin{equation}
    f_* \triangleq p(t_*|\mathcal{F}) \sim \mathcal{GP}(\mu_*, \sigma_*).
\end{equation}
Therefore, by learning a period of historical force pattern $\mathcal{F}$, we can shortly generate the predicted force pattern distribution $\mathcal{F}_* = \{(t_*, f_*) | t_* \in \mathbf{t}_*, f_* \sim \mathcal{GP}(\mu_*, \sigma_*)\}$ over the near future via GPR. Since the $\mathbf{t}$, $\mathbf{t_*}$ and $\mathbf{f}$ are all one-dimensional (1D) vectors, the computational burden from GPR is quite small, enabling the proposed method work in a real-time manner. 

To find the function $\xi$ in \prettyref{prob:proxi_prob}, we can determine the jamming state by monitoring the anomaly occurring in the force pattern, which is evaluated by the z-score between real force measurement $f_t$ and its prediction $f_*$ at time $t_*$: 
\begin{equation}\label{eq:z_scores}
        \mathbf{z} = \{(f_t|_{t = t_*} - \mu_*)/ \sigma_*\}, \forall t_* \in \mathbf{t}_*\ .
\end{equation}
When there is no object around the probe, the z-scores of new observations are expected to stay within a high confidence interval (CI). Normally, for the Gaussian distribution, the z-score for 99\% CI is 2.576. In contrast, a sudden and sharp increase in z-scores indicates the anomaly in the force pattern, probably resulting from a subterranean object in the vicinity of the probe. We online monitor the changes in the z-score, and once a certain threshold $\Bar{z}$ is exceeded, our system could release a proximity warning, followed by a stop command sent to the robot arm. Thus, we can mathematically express our haptic-based proximity sensing system as the solution of \prettyref{prob:proxi_prob} as follows:
\begin{equation} \label{eq:grains_eq}
    \xi (\mathcal{T}) =\left\{
    \begin{array}{rcl}
    0 &, & \ \ z_i < \Bar{z}, \forall z_i \in \mathbf{z}, \\
    1 &, & \ \  z_i \geq \Bar{z}, \exists z_i \in \mathbf{z}.\\
    \end{array} \right.
\end{equation}
It is obvious that the selection of the z-score threshold $\Bar{z}$ in \prettyref{eq:grains_eq} will be vital since it directly determines our tolerance for outliers and, in turn, when to stop moving to avoid the potential collision. In the following, we will introduce how to adaptively select system parameters (including $\Bar{z}$) for different particles through the proposed parameter calibration experiments.

\subsection{Generalization to Various GMs}
To adapt to different granular scenarios, we have developed an automated parameter calibration process, which can autonomously select the optimal parameter set $\mathcal{D}$ for the current GM, including $\mathcal{D} = \{\text{MV}^*, T^*, \Bar{z}\}$. Recall that the variable MV is a dimensionless parameter, ranging from $0$ to $1$, to control the speed of the end-effector of the robot arm, as introduced in \prettyref{sec:motion_control}. The hyperparameter $T$ used in \prettyref{eq:periodic_kernel} represents the periodicity of the force pattern $\mathcal{F}$. Even if $T$ will be tuned during the GPR learning, an optimal value $T^*$ as the initial value for the online learning will reduce the computational cost. Most importantly, the periodicity $T$ only relies on the parameters of spiral trajectory, i.e., CR, AV, MV, and be determined by the following equation:
\begin{equation}\label{eq:T_equation}
    T = \frac{l_p}{V_{max}\cdot \text{MV}} f_s,\ \text{MV} \in [0, 1],
\end{equation}
where $V_{max}$ is the max speed of the end-effector and $f_s$ is the sampling frequency of F/T sensor. In this work, $V_{max} = 0.08968$ m/s and $f_s = 62.5$ Hz according to the setup used. Also, $l_p$ refers to the length of the spiral path for one circular motion, as in \prettyref{fig:def_spiral_traj}. It can be calculated by the sum of segment-wise paths from discretized path points, i.e.,
\begin{align}\label{eq:path_length_one_spiral}
    l_p &= \sum_{i=0}^{H-1} || \mathbf{x}_i - \mathbf{x}_{i+1}||_2.
\end{align}
Given CR and AV, we can obtain the $l_p$ by substituting $\mathbf{x}_i, \mathbf{x}_{i+1}$ from \prettyref{eq:path_xi} into \prettyref{eq:path_length_one_spiral}. Then the optimal periodicity $T^*$ can be determined in advance by submitting $l_p$ and the given MV$^*$ into \prettyref{eq:T_equation}.
In addition, from our preliminary experiments, we find that different types of GMs display varying z-score thresholds to identify the jamming state. To increase credibility and reduce the ratio of false positive cases in various GMs, the GM-specific parameter $\Bar{z}$ should be calibrated in advance as well.

To do so, we take samples from current granules and store them in a container. Our GRAINS then controls the probe raking in this container along a spiral trajectory with given CR and AV, as well as MV$_i$ from the set $\{$MV$_i\}$. For each MV$_i$, GRAINS calculates the periodicity $T_i$ by \prettyref{eq:T_equation}. Then we can formulate an ESS kernel by substituting $T_i$ in \prettyref{eq:periodic_kernel}. By using the sum-kernel \prettyref{eq:sum_kernel} from \prettyref{eq:periodic_kernel} and \prettyref{eq:noise_kernel}, we can initialize a GP model from \prettyref{eq:def_GP}, denoted by $GP_i$. Note that other parameters in \prettyref{eq:periodic_kernel} and \prettyref{eq:noise_kernel} will be given fixed values, and they will be tuned by GPR learning.
At each MV$_i$, the GRAINS controls the probe to explore $20$ cm in the container.
After that, GRAINS successively divides the online measurements $\mathcal{T}$ into several force patterns and feeds the $k$-th force pattern $\mathcal{F}_k$ into $GP_i$. Based on the predicted mean and standard deviation from $GP_i$, GRAINS computes the z-scores of the $(k+1)$-th force pattern $\mathcal{F}_{k+1}$. This process is repeated to obtain z-scores for all force patterns except the first one. GRAINS then calculates the Root Mean Squared Error (RMSE) of z-scores at each MV$_i$, denoted as $\text{RMSE}_i$. Finally, GRAINS determines the optimal MV$^*$ corresponding to the minimum RMSE values. The periodicity prior $T^*$ at MV$^*$ is then chosen as the initial hyperparameter for \prettyref{eq:periodic_kernel}. In addition, since no object is located under GMs, the maximum z-score at MV$^*$ during this process is considered as the threshold $\Bar{z}$ for \prettyref{eq:grains_eq}.

So, it can be seen that the above calibration process is all automated, and optimal (hyper)parameters in $\mathcal{D}$ are all determined autonomously by the proposed system GRAINS according to the type of granules, without human factors. 

\section{Experiment Results}
\subsection{Proximity Sensing in Sands}
\begin{figure}[htbp]
    \centering
    \includegraphics[width=0.99\textwidth]{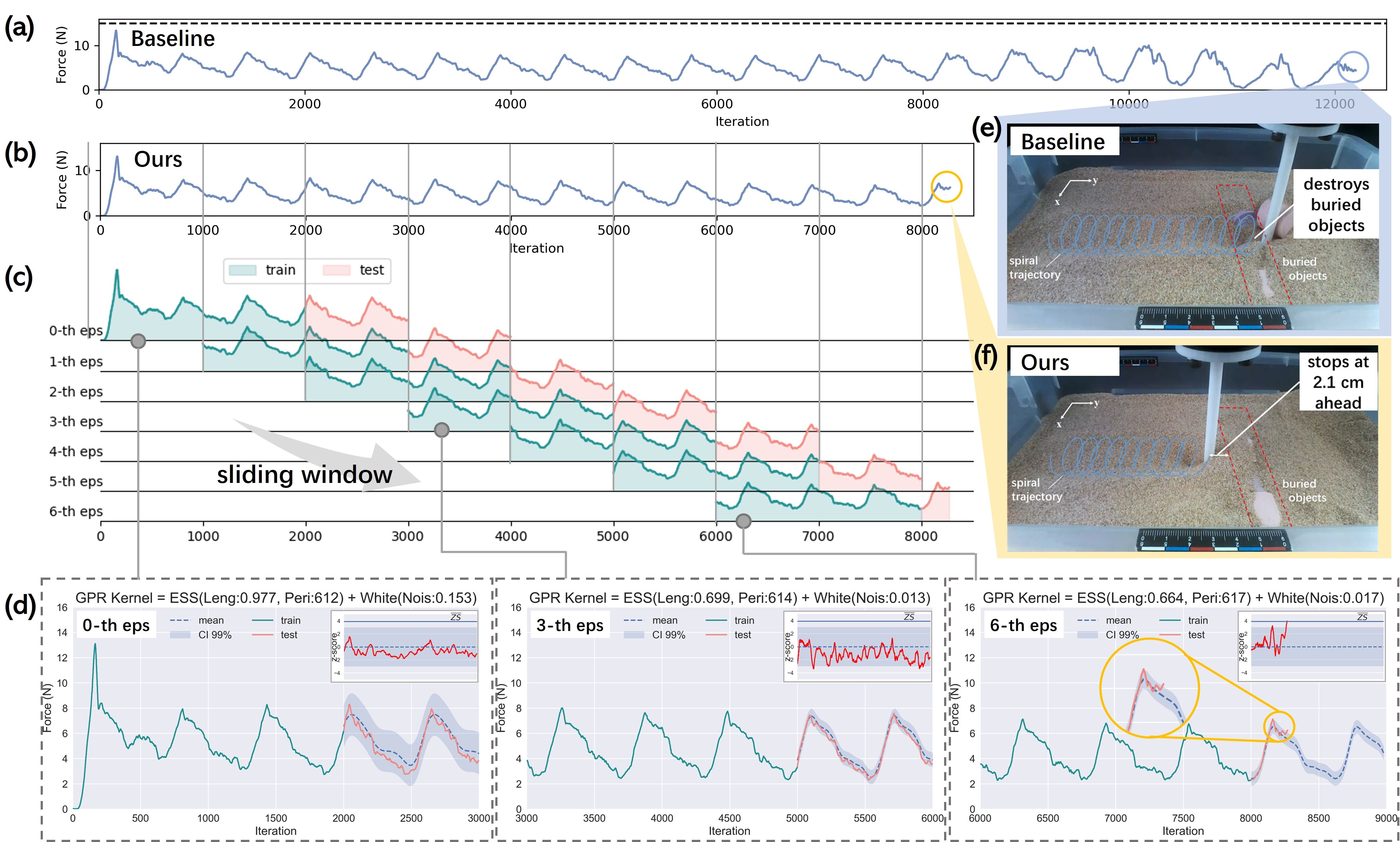}
    \vspace{-10pt}
    \caption{Proximity sensing experiments in sands. (a),(e) Baseline method. (b)-(d),(f) Our mehtod. }
    \label{fig:exp_results_sand}
    \vspace{-15pt}
\end{figure}

First of all, we demonstrate the near-field perception of the proposed proximity sensing system GRAINS in sands, as shown in \prettyref{fig:exp_results_sand}. Here the sand is composed of silica with diameters ranging from $0.5$ to $1.6$ mm.

To determine the optimal (hyper)parameters for sands, we set the speed MV$_i = \{0.2, 0.3, \cdots, 0.7\}$ and generate the spiral path by CR $= 0.02$ m and AV $=0.01$ m. 
These settings are based on our experimental observations. If MV is too small, low search efficiency is not acceptable. Also, MV cannot be too large; otherwise, it is too late for an emergency stop. Large CR and AV values may cause the exploration to exceed the existing sandbox space.
After the automatic calibration, the resulting RMSE values corresponding to MV$_i$ are revealed in \prettyref{tab:parameter_cali_sand}. Thus, we obtain the optimal (hyper)parameters as $\mathcal{D} = \{\text{MV}^*:  0.2, T^*: 439, \Bar{z}: 3.9\}$.
After that, we validate the proximity sensing performance in sands, where a wooden cylinder is buried, as depicted in \prettyref{fig:exp_results_sand}-(f). From the depth assumption in \prettyref{sec:problem_statement}, the penetration depth of the probe in the experiment is fixed at $4$ cm, and the underground objects are located at a relatively shallow depth, less than $4$ cm, to ensure the overlap in the depth direction between the probe and the buried objects. 
Before the real experiment, the F/T sensor will be zeroed out to prevent drift. During the experiment, the force sensor will use a low-pass filter to filter the original data to reduce the influence of noise. Then GRAINS monitors the filtered force measurements $f_t$ in a real-time manner and employs the sliding window to clip the force pattern $\mathcal{F}$ with $M = 2000$ and predict the force pattern distribution $\mathcal{F}_*$ over the next $M_* = 1000$ by \prettyref{eq:gpr_posterior} with pre-calibrated optimal parameters $\mathcal{D}$. The z-score at $t_* \in \mathbf{t}_*$  is constantly computed from \prettyref{eq:z_scores} once the force measurement $f_t|_{t=t_*}$ is available.  

The results of proximity sensing from GRAINS are illustrated in \prettyref{fig:exp_results_sand}-(f). The measurements $\mathcal{T}$ during the spiral trajectory can shown in \prettyref{fig:exp_results_sand}-(b). The window sliding operation is shown in \prettyref{fig:exp_results_sand}-(c). A window of $M = 2000$ iterations is sliding along $\mathcal{T}$ to form the $\mathcal{F}$ (green curves in \prettyref{fig:exp_results_sand}-(d)), and z-scores $\mathbf{z}$ of the next $M_* = 1000$ iterations are calculated accordingly, as illustrated in the insets of \prettyref{fig:exp_results_sand}-(d), by the real measurements $f_t|_{t\in \mathbf{t}_*}$  (pink curves in \prettyref{fig:exp_results_sand}-(d)) and predicted distribution $\mathcal{F}_*$ (blue dashed curve and blue shaded area in \prettyref{fig:exp_results_sand}-(d)).  
It can be seen that when the probe is far away from the object, e.g., $0$-th eps and $3$-th eps in \prettyref{fig:exp_results_sand}-(d), the predicted force pattern is consistent with the real measurements, as $f_t|_{t=t_*}$ is within CI $99\%$, or equivalently $z_i < \Bar{z} = 3.9,\ \forall z_i \in \mathbf{z}$. So, by definition of \prettyref{eq:grains_eq}, GRAINS safely outputs $\xi(\mathcal{T}) = 0$. When the probe approaches the object, the failure wedge zone contacts the object before the probe as expected, and granule jamming occurs. Thus, the force pattern induced by the jamming state starts to diverge from the high CI area with the growth of z-scores.
Once a z-score value $z_L$ exceeds the predetermined threshold $\Bar{z}$ at time $t_L$ during the $6$-th eps as shown in \prettyref{fig:exp_results_sand}-(d), GRAINS identifies the presence of subsurface stuff nearby, and outputs $\xi(\mathcal{T}|_{t=t_L}) = 1$. Therefore GRAINS sends a cease command to stop the robot arm at the current position $\mathbf{x}_t|_{t=t_L}$ to avoid the potential collision accordingly. From \prettyref{fig:exp_results_sand}-(f), the GRAINS perceives the presence of near objects beneath sands at a distance of $2.1$ cm in advance. From \prettyref{eq:sensing_range}, we can say the proximity sensing range for this case is $\zeta(\xi) = 2.1$ cm. 

We compare our method with the baseline approach, which uses the pre-defined threshold for the \textit{force value}, as in \cite{jia2022autonomous}. In this case, we set the force bar as $15$ N, as shown in the dashed line in \prettyref{fig:exp_results_sand}-(a), considering the maximum static friction force from still particles that the probe should overcome. From \prettyref{fig:exp_results_sand}-(e), we can observe that the probe fails to detect the proximity of underground objects and breaks into several pieces finally. In addition, if the force threshold is lower than $15$ N, then the probe could stuck at the initial stage since this value should be larger than the maximum static friction force, as depicted in \prettyref{fig:exp_results_sand}-(a). Therefore, the baseline method requires great care to determine the force threshold. And, for different granules, these human efforts need to be repeated. However, our method can autonomously update parameters to adapt to different GMs, as evaluated in the following.

\begin{table*}[tbp]
    \centering
    \caption{Parameter calibration for the sand. Bold for the minimum RMSE value.}
    \label{tab:parameter_cali_sand}
    \resizebox{\textwidth}{!}{%
    \begin{tabular}{@{}cccccccccccc@{}}
    \toprule
        \multicolumn{2}{c|}{MV$_1$($0.2$) / $T_1$($439$)} & \multicolumn{2}{c|}{MV$_2$($0.3$) / $T_2$($293$)} & \multicolumn{2}{c|}{MV$_3$($0.4$) / $T_3$($220$)} & \multicolumn{2}{c|}{MV$_4$($0.5$) / $T_4$($176$)} & \multicolumn{2}{c|}{MV$_5$($0.6$) / $T_5$($146$)} & \multicolumn{2}{c}{MV$_6$($0.7$) / $T_6$($125$)} \\
        \cmidrule(l){1-12}
        \multicolumn{1}{c}{RMSE} & \multicolumn{1}{c|}{max($\mathbf{z}$)} & \multicolumn{1}{c}{RMSE} & \multicolumn{1}{c|}{max($\mathbf{z}$)} & \multicolumn{1}{c}{RMSE} & \multicolumn{1}{c|}{max($\mathbf{z}$)} & \multicolumn{1}{c}{RMSE} & \multicolumn{1}{c|}{max($\mathbf{z}$)} & \multicolumn{1}{c}{RMSE} & \multicolumn{1}{c|}{max($\mathbf{z}$)} & \multicolumn{1}{c}{RMSE} & \multicolumn{1}{c}{max($\mathbf{z}$)} \\ \midrule\midrule
        $\mathbf{0.9191}$ & ${3.9}$ & $1.3739$ & $7.0$ & $1.4695$ & $8.7$ & $1.2284$ & $4.2$ & $1.9801$ & $12.4$ & $1.2498$ & $4.3$\\
    \bottomrule
    \end{tabular}
    }
    \vspace{-15pt}
\end{table*}

\subsection{Proximity Sensing in Various GMs}
\begin{figure}[htbp]
    \centering
    \vspace{-20pt}
    \includegraphics[width=0.90\textwidth]{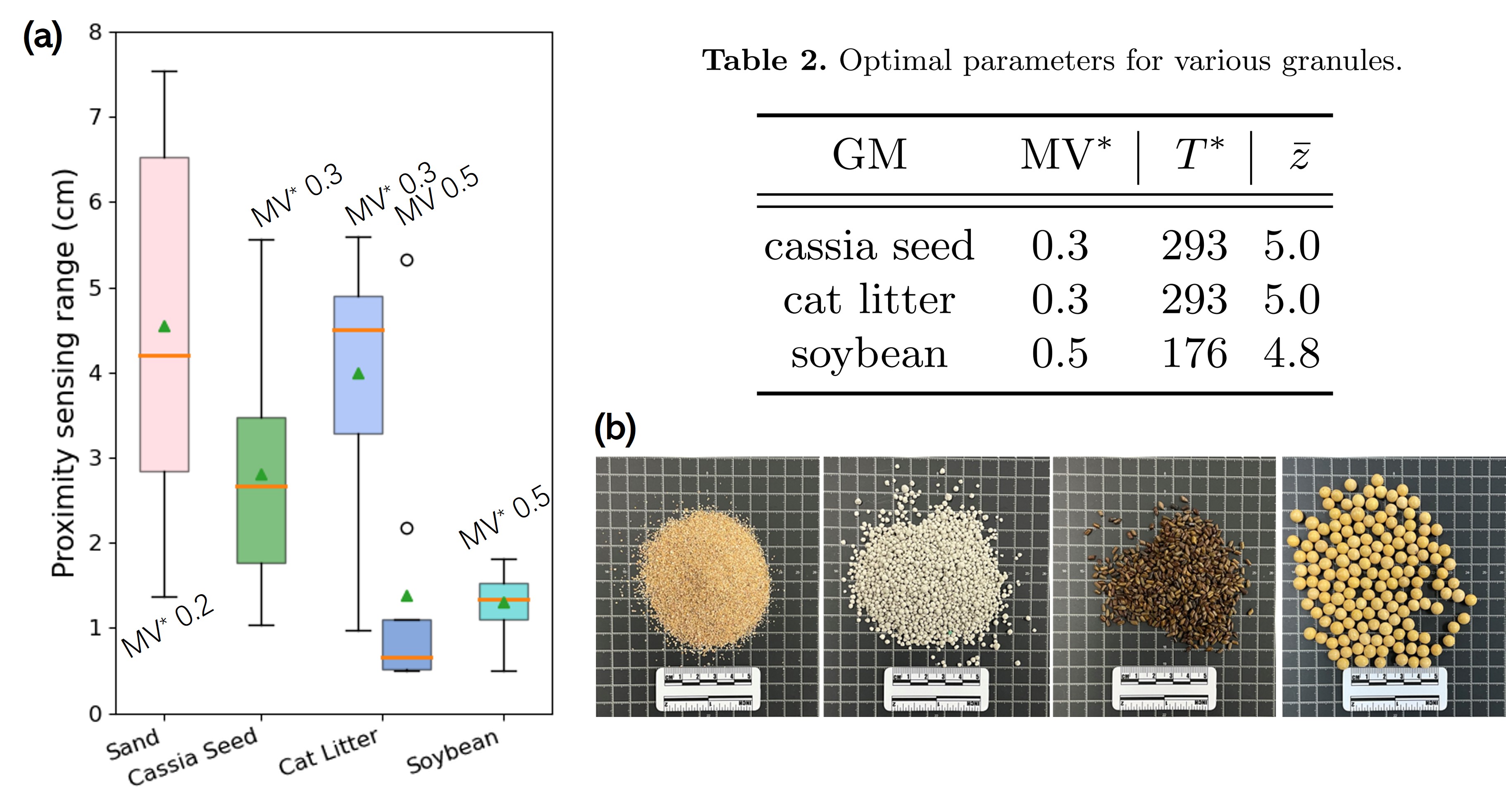}
    \vspace{-15pt}
    \caption{Proximity sensing experiments in various GMs. (a) The distribution of proximity sensing range $\zeta$. (b) Tested granules in this work. The scale bar represents $5$ cm.}
    \label{fig:exp_results_others}
\end{figure}

We test our GRAINS in other granules, as shown in \prettyref{fig:exp_results_others}-(b), including cassia seed (about $3.5$ mm in length and $2.5$ mm in width), cat litter (about $3 \sim 4$ mm in diameter), and soybean (about $6.5$ mm in diameter). Similar to the calibration process for the sand, the optimal parameters for these GMs are determined by GRAINS autonomously and presented in Table 2. Then we conduct proximity perception in each GM for $20$ times and resulting proximity sensing ranges $\zeta(\xi)$ defined by \prettyref{eq:sensing_range} are recorded. From \prettyref{fig:exp_results_others}-(a), we can find that the median value of $\zeta(\xi)$ in cat litter ($\sim 4.5$ cm) is similar to that in the sand ($\sim 4.2$ cm) but with a smaller dispersion. This discrepancy can potentially be attributed to the notably larger particulate size of cat litter, which leads to increased instability in the conduction of force chains compared to sand. 
Since a rough granule surface is beneficial for forming a highly compressed state, the smoother surface of cassia seed requires the probe to get closer to objects to sense force anomaly signals from granular jamming, resulting in a shorter $\zeta(\xi)$ ($\sim 2.7$ cm), even if cassia seed has a similar grain size to cat litter. Soybean has the shortest $\zeta(\xi)$ ($\sim 1.3$ cm) due to its largest particle size and smoothest surface. 

In addition, we test the effects of non-optimal parameters on $\zeta(\xi)$. As shown in \prettyref{fig:exp_results_others}-(a) about the cat litter, the probe stops noticeably closer to the object with a non-optimal MV$=0.5$ than that with the optimal MV$^*=0.3$, i.e., $0.6$ cm and $4.5$ cm, respectively. Furthermore, several outliers can be identified in the case of MV $0.5$. This highlights the necessity and importance of parameter calibration in GRAINS.


\section{Conclusion}
In this paper, we propose a haptic-based proximity sensing system, GRAINS, for buried objects in granular material. We leverage the failure wedge zone, the unique characteristic of particles, when a probe moves in GMs as the near-filed perception medium. We employ GPR to online learn the force pattern from the haptic feedback of the probe-GM interaction, and then monitor the pattern anomaly resulting from the granule jamming when the failure wedge zone contacts the object. Therefore, we use the failure wedge zone ahead of the probe as an airbag and avoid the probe-object collision by recognizing the contact between the failure wedge zone and the object in advance, so as to achieve proximity perception. We design a novel spiral trajectory to control the failure wedge zone, to enlarge the sensing area. In this system, a parameter calibration process is introduced, so that the GRAINS can autonomously update parameters and robustly work for different granules. The limitation of current GRAINS may be that it can only operate within single homogeneous particles at present. The deployment of GRAINS in real complex particles is a future work. In addition, proximity sensing in 3D space is a promising direction.
%
%
\bibliographystyle{splncs04}
\bibliography{mybibfiles}

\end{document}